# Information Forests

Zhao Yi    Stefano Soatto    Maneesh Dewan    Yiqiang Zhan

University of California, Los Angeles, {`zyi,soatto`}@cs.ucla.edu
Siemens Medical Solutions, {`first.last`}@siemens.com

## Abstract

We describe Information Forests, an approach to classification that generalizes Random Forests by replacing the splitting criterion of non-leaf nodes from a discriminative one – based on the entropy of the label distribution – to a generative one – based on maximizing the information divergence between the class-conditional distributions in the resulting partitions. The basic idea consists of deferring classification until a measure of "classification confidence" is sufficiently high, and instead breaking down the data so as to maximize this measure. In an alternative interpretation, Information Forests attempt to partition the data into subsets that are "as informative as possible" for the purpose of the task, which is to classify the data. Classification confidence, or informative content of the subsets, is quantified by the Information Divergence. Our approach relates to active learning, semi-supervised learning, mixed generative/discriminative learning.

## 1 Introduction

We introduce Information Forests (IFs), a family of part-based classifiers designed for problems that are not easily solvable as a whole. In IFs there is a hidden location or selection variable that is key to performing classification: While there may be no distinguishing characteristic between the positive and negative samples considered as a whole, one can find "informative subsets" (regions, parts, or groups) where classification is simple to carry out. However, IFs are not restricted to these problems, and can be interpreted as a generic family of classifiers that includes Random Forests (RFs) as a special case.

The motivation comes from problems such as detection of people in images, where the distribution of intensity or color values in the region occupied by a person is not discriminative, and could be identical to the distribution of intensity or color values outside the same region. However, when the problem is restricted to smaller regions, or "parts," the problem may be more easily solved.

## 1.1 Intuition

The key idea of Information Forests is to defer attempts to *classify* data points, and focus first on *grouping* them in a way that makes classification as simple as possible. In other words, the goal at the outset is *not* to partition the data into clusters that are as *"pure"* as possible (belonging to the same class). Instead, the goal is to partition the data into clusters that are *as simple as possible to classify* down the line, and only perform the classification when it becomes sufficiently simple. In other words yet, the focus is to break down the original classification problem (for the entire dataset) into smaller subsets that are as simple as possible to classify. Only when the classification problem is "simple enough" it is actually carried out. Otherwise, the grouping process proceeds in a recursive, hierarchical fashion. In this *divide-et-impera* scheme, the goal is to determine groups of data that are *as informative as possible* for the purpose of the task, which is the determination of the class label $\lambda$. Such groups can be considered "regions" or "parts" or "subsets" depending on the application. This is illustrated in Fig. 1

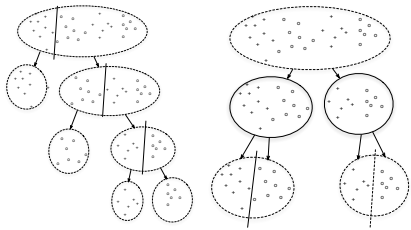

Figure 1: *Random Forest vs. Information Forest. A sequence of n groups alternating positive/negative/positive/negative etc. partitioned using a Random Forests with linear stumps requires a number of levels that grows linearly with n (left). An Information Forest using the same stumps (right) does not try to classify samples immediately, but instead tries to partition them into groups that are simple to classify, and defers the decision until confidence $\tau$ is sufficiently high and information gain $\delta$ sufficiently small.*

## 1.2 Formalization

Let $\lambda \in \{0,1\}$ be a binary class label, $x \in D \subset \mathbb{R}^k$, with $k = 2, 3$ a location variable, and $y : D \to Y$, $x \mapsto y(x)$ a measurement (or "feature") associate to location $x$, that takes values in some vector space $Y$. When the domain $D$ is discretized (e.g., the planar lattice), $x$ can be identified with an index $i \in \Lambda \mid x_i \in D$. In that case, we indicate $y(x)$ simply by $y_i$.

A (binary[1]) *segmentation problem* consists of partitioning the spatial domain

---
[1]Extension to multi-class segmentation, where $\lambda \in \{1, 2, \ldots, M\}$ is straightforward and will therefore not be considered here.

$D$ into two regions, $\Omega$ and $D\backslash\Omega$, according to the value of the feature $y(x)$. This can be done by considering the posterior probability

$$P(\lambda|y) \propto p(y|\lambda)P(\lambda), \qquad (1)$$

where the first term on the right hand side indicates the likelihood, and the second term the location prior. It should be clear that meaningfully solving this problem hinges on the two likelihoods, $p(y|\lambda = 1)$ and $p(y|\lambda = 0)$ being different:

$$p(y|\lambda = 1) \neq p(y|\lambda = 0). \qquad (2)$$

If this is the case, we can infer $\lambda$ and, from it, $\Omega = \{x \mid \lambda(x) = 1\}$. However, there are plenty of examples where where (2) is violated. We refer to problems where the condition (2) is violated as problems that *"are not solvable as whole"*, in the sense that we cannot segment the spatial domain simply by comparing statistics inside $\Omega$ to statistics outside. Nevertheless, it may be possible to determine *parts*, or local regions $S_i \subset D$, within which the likelihoods are different:

$$\exists \{S_j\}_{j=1}^N \mid p(y|x \in S_j, \lambda = 1) \neq p(y|x \in S_j, \lambda = 0),$$
$$S_j \subset D, \; j = 1, \ldots, N. \qquad (3)$$

Note that the collection $\{S_j\}$ is not unique, does not need to form a partition of $D$, as there is no requirement that $S_i \cap S_j \neq \emptyset$ for $i \neq j$, so long as the union of these regions cover[2] $D$. The regions $S_j$ do not even need to be simply connected. In some applications, one may want to impose these further conditions.

In the discrete-domain case, we identify the index $i$ with the location $x_i$, so the regions become subsets of the data. With an abuse of notation, we write

$$S_j = \{i_1, i_2, \ldots, i_{n_j}\}. \qquad (4)$$

Therefore, we write the two conditions (2)-(3) as

$$\boxed{p(y_i|\lambda_i = 1) = p(y_i|\lambda_i = 0),}$$
$$\boxed{p(y_i|i \in S_j, \lambda_i = 1) \neq p(y_i|i \in S_j, \lambda_i = 0).} \qquad (5)$$

Assuming these conditions are satisfied, we can write the posteriors by marginalizing over the sets $S_j$,

$$p(\lambda|y_i) \propto \sum_j p(y_i \mid i \in S_j, \lambda)P(i \in S_j|\lambda)P(\lambda) \qquad (6)$$

or by maximizing over all possible collections of sets $\{S_j\}$. In either case, the sets $S_j$ are not known, so the segmentation problem is naturally broken down into two components: One is to determine the sets $S_j$, the other is to determine the class labels within each of them:

---

[2]Indeed, even this condition can be relaxed to assuming that these regions cover the boundary of $\Omega$, $\cup_j S_j \supset \partial\Omega$, by making suitable assumptions on the prior $p(\lambda|x)$.

**Given** a training set of labeled samples $\{y_i, \lambda_i\}_{i=1}^M$,

**Find** a collection of sets $\{S_j\}_{j=1}^N$ such that $S_j \subset D$ and $D \subset \cup_j S_j$, that are "as informative as possible" for the purpose of determining the class label $\lambda$.

**If** the sets are "sufficiently informative" of $\Omega$, perform the classification; that is, determine the label $\lambda$ within these sets.

The key condition translates to the restricted likelihoods $p(y_i | i \in S_j, \lambda = 1)$ and $p(y_i | i \in S_j, \lambda = 1)$ being "as different as possible" in the sense of relative entropy (information divergence, of Kullback-Liebler divergence). When they are sufficiently different, the set is sufficiently informative of $\Omega$, and classification can be easily performed by comparing likelihood or posterior ratios.

This problem relates to *active learning*, in the sense that the classifier has to *select*, among all possible subsets, the ones that are *informative* in the sense of enabling the classification $\lambda$. A possible approach would be to select $S_i$ at random. However, an active learner would want to choose, among all possible $S_i$, the ones that are *most informative* towards solving the original classification problem, that is to determine $\lambda$. It also relates to *semi-supervised learning* with model selection, since – in addition to determining the discrete variable $\lambda$ for which supervision is provided via the training set – one has to determine the sets $S_j$, that can be interpreted as groupings, or collections, or subsets of the training data. However, no supervision is given as to which point $x \in D$ belongs to which group $S_i$. In addition, the number of such regions $N$ is not known and has to be inferred (model selection). This problem also touches on the issue of generative/discriminative models, since the groups $S_j$ can be interpreted as generative (latent mixture model), while the ultimate goal is classification.

Information Forests implement the program above using the machinery of boosting and decision trees, as we describe next.

## 2   Derivation of Information Forests

Information Forests are a family of classifiers that accomplish the goals described in the previous sections using the tools of randomized trees.

The groups ("clusters", or "regions") $S_j \subset D$ are chosen within a class $\mathcal{S}$ defined by a family of simple classifiers (decision stumps). For convenience, we expand the index $j$ into two indices, one relating to the "features" $f_j$ and one relating to a threshold $\theta_k$. We then define, for a continuous location parameter $x$

$$S_{jk} \;\dot{=}\; \{x \in D \mid f_j(x, y) \geq \theta_k\} \tag{7}$$

where the feature $f : D \times Y \to \mathbb{R}$; $(x, y) \mapsto f(x, y)$ is any scalar-valued statistic and the threshold $\theta \in \mathbb{R}$ is chosen within a finite set. We call the set of features $\mathcal{F} \doteq \{f_j\}$ and the set of thresholds $\Theta = \{\theta_k\}$. The complement of $S_{jk}$ in $D$ is indicated with $S_{jk}^c \doteq \{x \in D \mid f_j(x, y) < \theta_k\} = D \backslash S_{jk}$. In the simplest case, for a grayscale image, we could have $f(x, y) = y(x)$ where $y(x)$ is the intensity value at pixel $x$. More in general, $f$ can be any (scalar) function of $y$ in a neighborhood of $x$. For the discrete case, where $i$ is identified with the location $x_i$, with an abuse of notation we write

$$S_{jk} = \{i \in \Lambda \mid f_j(y_i) \geq \theta_k\} \tag{8}$$

and again $S_{jk}^c = \{i \in \Lambda \mid f_j(y_i) < \theta_k\}$. Here the features $f$ are $f : \Lambda \times Y \to \mathbb{R}$; $(i, y) \mapsto f(y_i)$. Specifying the feature and threshold $(f_j, \theta_k)$ is equivalent to specifying the set $S_{jk}$ and its complement $S_{jk}^c$.

We are interested in building informative sets using recursive binary partitions, so at each stage we only select one pair $\{S_{jk}, S_{jk}^c\}$. Among all features in $\mathcal{F}$ and thresholds in $\Theta$, Information Forests choose the one that makes the set $S_{jk}$ "as informative as possible" for the purpose of classification. From (5) it can be seen that the quantity that measures the "information content" of a set $S_{jk}$ (or a feature $f_j, \theta_k$) for the purpose of classification is the Information Divergence (Relative Entropy, or Kullback-Liebler Divergence) between the distributions $p(y_i|i \in S_{jk}, \lambda_i = 1)$ and $p(y_i|i \in S_{jk}, \lambda_i = 0)$. In short-hand, we write $p(y_i|\cdots, \lambda_i = 1)$ as $p_1(y_i|\cdots)$ and $p(y_i|\cdots, \lambda_i = 0)$ as $p_0(y_i|\cdots)$ and

$$KL(f_j, \theta_k) = \frac{|S|}{|D|} \mathbb{KL}(p_1(y_i|i \in S) \parallel p_0(y_i|i \in S)) +$$
$$+ \frac{|S^c|}{|D|} \mathbb{KL}(p_1(y_i|i \in S^c) \parallel p_0(y_i|i \in S^c)). \tag{9}$$

From the characterization of the sets $S_{jk}$, $i \in S_{jk}$ is equivalent to $f_j(y_i) \geq \theta_k$, so we write $S_{jk} = S(f_j, \theta_k)$. Therefore, a decision stump ("KL-node") chooses among features and thresholds one (of the possibly many) that

$$\hat{f}_j, \hat{\theta}_k \doteq \arg\max_{f_j, \theta_k} \frac{|S(f_j, \theta_k)|}{|D|}$$
$$\mathbb{KL}\left(p_1(y_i|f_j \geq \theta_k) \| p_0(y_i|f_j \geq \theta_k)\right)$$
$$+ \frac{|S^c(f_j, \theta_k)|}{|D|} \mathbb{KL}\left(p_1(y_i|f_j < \theta_k) \| p(y_i|f_j < \theta_k)\right). \tag{10}$$

Here $\mathbb{KL}(p\|q) = \mathbb{E}_p\left[\ln \frac{p}{q}\right] = \int \ln \frac{p}{q} dP$ denotes the Kullback-Liebler divergence.[3] The normalization factors $|S|/|D|$ and $|S^c|/|D|$ count the cardinality of the set $S$ and its complement relative to the size of the domain $D$.

If the divergence value is sufficiently large, $KL(f_j, \theta_k) > \tau$, the positive and negative distributions are sufficiently different, and therefore the classification problem is easily solvable. To actually solve it, one could use the same decision stumps (features) $\mathcal{F}$, but now chosen to minimize the entropy of the distribution

---

[3]Several alternate divergence measures can be employed instead of Kullback-Leibler's, for instance symmetrized versions of it, or more general Jeffrey divergence.

of class labels, $p(\lambda_i | i \in S_{jk}) = p(\lambda_i | f_j \geq \theta_k)$, and its complement:

$$H(f_j, \theta_k) \doteq \frac{|S(f_j, \theta_k)|}{|D|} \mathbb{H}(\lambda_i | f_j \geq \theta_k) +$$
$$+ \frac{|S^c(f_j, \theta_k)|}{|D|} \mathbb{H}(\lambda_i | f_j < \theta_k) \quad (11)$$

where $\mathbb{H}(p) = \mathbb{E}_p[\ln p] = \int \ln p \, dP$ is the entropy of the distribution $p$. If the quantity (10) is sufficiently large, $\mathbb{KL}(f_j, \theta_k) > \tau$, (11) can be solved. If not, the process can be iterated, and the data further split according to the same criterion, the maximization of $KL(f_j, \theta_k)$. The value $\tau$ can therefore be interpreted as measuring the *least tolerable confidence* in the classification.

## 2.1 Implementation

Information Forests perform hierarchical grouping (mixture modeling) and classification by recursive binary partitioning. During **training**, starting from a the entire dataset $\{1, \ldots, N\}$, each node $S$ is passed through a Divergence Test:

$$\mathbb{KL}(p_1(y_i | i \in S) \parallel p_0(y_i | i \in S)) > \tau. \quad (12)$$

If this condition is satisfied, the node is designated as an H-node that solves

$$\boxed{\hat{f}_j, \hat{\theta}_k = \arg \min_{f \in \mathcal{F}, \theta \in \Theta} H(f, \theta)} \quad (13)$$

If the Information Gain is below a minimum threshold $\delta > 0$,

$$\mathbb{H}(\lambda_i | i \in S) - H(\hat{f}_j, \hat{\theta}_k) \leq \delta, \quad (14)$$

the node is re-designated as a terminal node ("leaf") and the classes are determined via

$$\hat{\lambda} = \arg \max_{\lambda_i \in \{0,1\}} p(\lambda_i | i \in S). \quad (15)$$

If the condition (12) is violated, the two classes are difficult to separate, so we look to partition the data into new clusters via a KL-node that solves

$$\boxed{\hat{f}_j, \hat{\theta}_k = \arg \max_{f, \theta \in \mathcal{F}} KL(f, \theta)} \quad (16)$$

In either case, so long as the node is not a leaf, the selected $\hat{f}_j, \hat{\theta}_k$ generates two sets, $S(\hat{f}_j, \hat{\theta}_k)$ and its complement, where

$$S(\hat{f}_j, \hat{\theta}_k) = \{i \in S \,|\, \hat{f}_j(y_i) \geq \hat{\theta}_k\}. \quad (17)$$

The two sets $S = S(\hat{f}_j, \hat{\theta}_k)$ and $S = S^c(\hat{f}_j, \hat{\theta}_k)$ are fed each to one of the two children of the current node as the tree grows. Like in a Random Forest, the process is repeated multiple times, for random subsets of the data points. During **testing**, each datum $y_i$ is run through the cascade of tests $\hat{f}_j(y_i) \geq \hat{\theta}_k$, on multiple trees, and then voting is performed.

## 2.2 Approximation and lower bound

While testing consists of repeated scalar tests that have trivial computational complexity, training requires multiple iterations of exhaustive optimization at each node, where each step entails computing $KL(f, \theta)$, that is a relative entropy between distributions in high-dimensional space (the feature space $Y$). Therefore, efficient approximations are needed.

One could employ several proxies of relative entropy, including Fisher scores. Or, one could compute relative entropy between scalar components (projections) of feature space. We approximate the Information Divergence with a lower bound

$$\mathbb{KL}(p_1(y_i|f_j \geq \theta_j) \parallel p_0(y_i|f_j \geq \theta_j)) \geq$$
$$\geq \mathbb{KL}(p_1(\Pi(y_i)|f_j \geq \theta_j) \parallel p_0(\Pi(y_i)|f_j \geq \theta_j)) \quad (18)$$

where $\Pi(y_i)$ is any 1-D projection of $y_i$. For ease of computation, we choose $\Pi(y_i) = f(y_i)$ from our feature pool. Since the previous inequality holds for any $\Pi$, we have

$$\mathbb{KL}(p_1(y_i|f_j \geq \theta_j) \parallel p_0(y_i|f_j \geq \theta_j)) \geq$$
$$\geq \max_{f \in \mathcal{F}, \theta} \mathbb{KL}(p_1(f(y_i)|f_j \geq \theta_j) \parallel p_0(f(y_i)|f_j \geq \theta_j)). \quad (19)$$

This process is repeated according to the same schedule of conventional Random Forests.

## 2.3 Analysis

Information Forests are a superset of Random Forest, as the former reduces to the latter when $\tau = 0$ is chosen. While it has been argued [1] that RF produce balanced trees, this is true only when the class $\mathcal{F}$ is infinite. In practice, $\mathcal{F}$ is always finite, and typically RFs produce heavily unbalanced trees, as the example in Fig. 1 illustrates. That example also shows that, when the dataset is not separable by the class of decision stumps, IFs produce more balanced and shallower trees when the set of classifiers is restricted.

More thorough analysis of the properties of IFs and the class of problems they are well matched to solve is forthcoming.

# 3 Discussion

Random Forests as a boosting variety of randomized decision trees, have been employed with a variety of splitting criteria, mostly related to entropy of the label distributions or mutual information between the features and the labels [5, 6, 2]. Breiman analyzes some of the properties of entropy and compares it with the Gini index in [1]. However, to the best of our knowledge, all of these approaches choose *discriminative splitting criteria*, where the goal is to produce partitions that are as pure as possible at each node, and there is no differentiation between leaf nodes and non-leaf nodes.

Several choices of decision stumps have also been applied, mostly depending on the application, with the simplest choices consisting of linear classifiers [3]. We have used simple linear scalar stumps for simplicity, but there is nothing in the derivation of IFs that precludes the use of more complex classifiers (other than computational considerations).

Since our approach mixes divergence measures and classification measures, the analysis of Nguyen et al. [4] could shed some light on the properties of the scheme proposed.

In forthcoming work, we intend to characterize the performance of IFs both empirically, as well as analytically.

## Acknowledgments

This project started in the summer of 2009 when Z. Yi was an intern at Siemens Medical Solutions. We wish to think Dr. Gerardo Hermosillo-Valadez for discussion during that phase. The continuation of this research was sponsored by DARPA under the MSEE program FA8650-11-1-7156, and by ARO under a MURI program W911NF-11-1-0391.

## References

[1] L. Breiman. Technical note: Some properties of splitting criteria. *Machine Learning*, 24(1):41–47, 1996.

[2] R. M. Goodman and P. Smyth. Decision tree design using information theory. *Knowledge Acquisition*, 2(1):1–19, 1990.

[3] T. K. Ho. The random subspace method for constructing decision forests. *IEEE Trans. Pattern Anal. Mach. Intell.*, 20(8):832–844, 1998.

[4] X. L. Nguyen, M. J. Wainwright, and Jordan M. I. On surrogate loss functions and f-divergences. *The Annals of Statistics*, 37(2):876–904, 2009.

[5] I. K. Sethi and G. P. R. Sarvarayudu. Hierarchical classifier design using mutual information. *IEEE Trans. Pattern Anal. Mach. Intell.*, (4):441–445, 1982.

[6] Q. R. Wang and C. Y. Suen. Analysis and design of a decision tree based on entropy reduction and its application to large character set recognition. *IEEE Trans. Pattern Anal. Mach. Intell.*, (4):406–417, 1984.